\pdfoutput=1

\documentclass[11pt]{article}

\usepackage[preprint]{acl}

\usepackage{times}
\usepackage{latexsym}

\usepackage[T1]{fontenc}

\usepackage[utf8]{inputenc}

\usepackage{microtype}

\usepackage{inconsolata}

\usepackage{graphicx}

\usepackage{amsthm,amsmath,amssymb}
\usepackage{bbding}
\usepackage{float}
\usepackage{bm}
\usepackage{booktabs}
\usepackage{graphicx}
\usepackage{multirow}
\usepackage{diagbox}
\usepackage{enumitem}
\usepackage{setspace}
\usepackage{algorithm}
\usepackage{algorithmic}
\usepackage{soul}
\usepackage{xspace}

\title{Investigating More Explainable and Partition-Free Compositionality\\Estimation for LLMs: A Rule-Generation Perspective}

\author{
Ziyao Xu$^1$, Cong Wang$^2$, Houfeng Wang$^1$\\
$^1$National Key Laboratory for Multimedia Information Processing, \\
School of Computer Science, Peking University \\
$^2$OPPO AI Center \\
\texttt{\{xzyxzy,wanghf\}@pku.edu.cn}
}

\begin{document}
\maketitle
\begin{abstract}
Compositional generalization tests are often used to estimate the compositionality of LLMs. However, such tests have the following limitations: (1) they only focus on the output results without considering LLMs' understanding of sample compositionality, resulting in explainability defects; (2) they rely on dataset partition to form the test set with combinations unseen in the training set, suffering from combination leakage issues. In this work, we propose a novel rule-generation perspective for compositionality estimation for LLMs. It requires LLMs to generate a program as rules for dataset mapping and provides estimates of the compositionality of LLMs using complexity-based theory. The perspective addresses the limitations of compositional generalization tests and provides a new way to analyze the compositionality characterization of LLMs. We conduct experiments and analysis of existing advanced LLMs based on this perspective on a string-to-grid task, and find various compositionality characterizations and compositionality deficiencies exhibited by LLMs. Our code is available at \url{https://github.com/xzy-xzy/RGP}.

\end{abstract}

\section{Introduction}

Compositionality is a concept that originates in the philosophy of language. It is a property that a language has to a certain extent and can be expressed as "the meaning of a complex expression is determined by its structure and the meanings of its constituents" \cite{pelletier1994principle, JANSSEN1997417, sep-compositionality, Compositionality-DaV}. In machine learning, the concept of compositionality is generalized to the mapping of inputs to outputs, suggesting that the output is determined by the meanings of the components of the input and the form in which the components are combined \cite{SCAN, PCFG}. In the NLP domain, many tasks involve mappings with significant compositionality, such as semantic parsing \cite{CFQ}, data-to-text generation \cite{SPOR}, compositional reasoning \cite{compositional-reasoning}, etc.

For a task that involves mappings with compositionality, if a model can recognize the compositionality of the mappings and utilize it, then the model will be able to correctly map the inputs made up of components to the outputs, as long as it knows the meaning of the components. This ability to recognize the compositionality of the mappings and utilize it is called the model's compositionality. %
Models' compositionality characterizes an effective form of reaching out-of-distribution generalization \cite{Systematic-Generalization} and this form is typical in human intelligence \cite{Symbols-and-mental-programs}. Therefore, the compositionality of models is an important research topic from both practical and cognitive perspectives \cite{summary-of-generalization}.

The research on models' compositionality has long been controversial, and the controversy focuses on how to properly measure a model's compositionality and whether the existing paradigms enable models to develop sufficient compositionality. In the NLP domain, a widely used approach to study the compositionality of language models on specific tasks is to conduct compositional generalization tests. The essence of the compositional generalization test is to partition the training and test sets with compositional differences, and then test the trained model's performance on the test set. After the emergence of large language models (LLMs), compositional generalization tests are still widely used under in-context learning for LLMs that are difficult to fine-tune directly.

The results of the compositional generalization test are intuitively suitable as a reflection of the compositionality of LLMs. However, compositional generalization tests have the following limitations: (1) they only focus on the output results with a gap between the LLMs' understanding of sample compositionality, leading to deficiencies in explainability; (2) they rely on dataset partition and suffer from potential composition leakage issues. The limitations make it difficult to obtain convincing estimates and analyses of the LLMs' compositionality, hindering more in-depth research on the compositionality of LLMs.

To solve this problem, we propose a rule-generation perspective for compositionality estimation for LLMs. In this perspective, we require LLMs to generate a program that describe the generation rules of a dataset, and use complexity-based theory to estimate the compositionality of LLMs based on the rules. By rule generation and appropriately quantifying the compositionality reflected in the rules, this perspective solves the limitations of compositional generalization tests. The perspective is consistent with intuitions about compositionality and generalization, and provides new ways to characterize the compositionality of LLMs.

Based on this perspective, we experiment and analyze advanced LLMs including reasoning models and non-reasoning models on a simple string-to-grid task. We identify different compositionality characterizations exhibited by LLMs, and compositionality defects of LLMs in various situations.

\section{Compositional Generalization Tests}
\begin{figure*}[t]
    \centering
    \includegraphics[width=\textwidth]{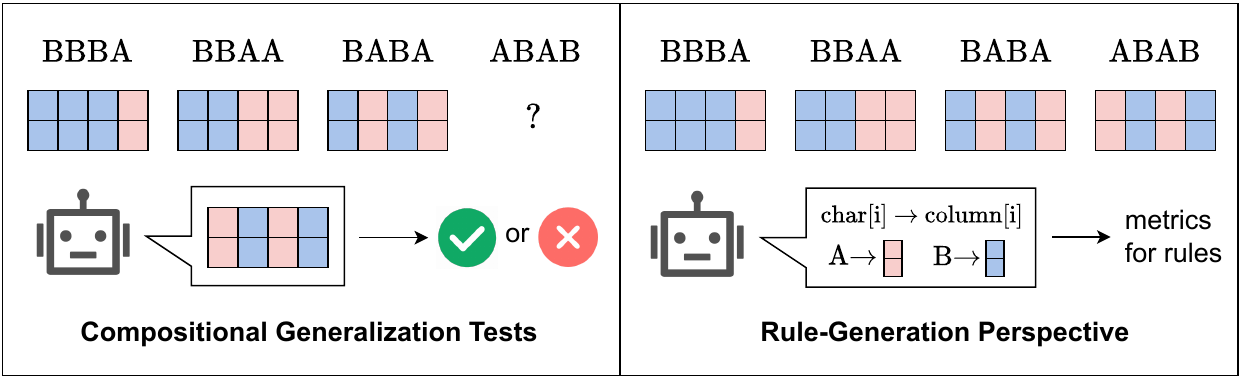}
    \caption{Illustrations of compositional generalization tests and rule-generation perspective on a string-to-grid task. Compositional generalization tests (left) partition the dataset and verify whether the model's output is correct for combinations not seen in the training set. The rule generation perspective (right) requires the model to generate rules describing the dataset and provide compositionality estimation based on those rules.}
    \label{fig:illustration}
\end{figure*}

In this section, we introduce the formulation of compositional functions and compositional generalization tests. We discuss the limitations of compositional generalization tests.

\subsection{Formulation}
Following the formulation in \citet{NEURIPS2023_15f6a108}, a compositional function $f$ transforms $K$ independent input components into $K$ output components, and then combines these output components into an output.  Formally, the $K$ independent input components are $K$ sets $C_1,...,C_K$, where $C_k = \{v_{k,1},...,v_{k,U}\}$ denotes the $U$ possible values of the $k$-th input component. For a value $c_k \in C_k$, the transformation function $\phi_k:C_k \rightarrow R_k$ transforms it into the output component $r_k$. The combination function $g: R_1 \times \cdots \times R_K \rightarrow Y$ combines the components into the output $y$. We define $X=C_1 \times \cdots \times C_K$ to denote the set containing all possible inputs. Given the input $x=(c_1, ..., c_K) \in X$, the compositional function $f:X \rightarrow Y$ can be expressed as:
\begin{equation}
    f(x) = g(\phi_1(c_1),...,\phi_K(c_K))
\end{equation}

For an unknown compositional function $f$, compositional generalization requires that the model be able to map unseen combinations of component values to expected outputs after seeing all the component values and some combinations of component values mapped to the outputs. Compositional generalization tests typically follow the training-test paradigm. In this paradigm, we partition $X$ into two disjoint subsets $X_S$ and $X_T$ that satisfy $\forall \ v_{k,j} \ , \ \ \exists \ x \in X_S \ , \ x_k = v_{k,j}$, and generate training set $S = \{(x, f(x)) \ | \ x \in X_S \}$ and test set $T = \{(x, f(x)) \ | \ x \in X_T \}$. The partition is usually based on minimizing the degree to which combinations of components in $T$ are visible in $S$ \cite{CFQ, COGS}. After a model is trained on the training set $S$, the model's accuracy on the test set $T$ is used to measure the model's compositional generalization performance. For LLMs that are difficult to fine-tune directly, each test of $x\in X_T$ is usually performed independently by extracting a subset of $S$ that covers the values in $x$ to be input to the LLMs as a demonstration of in-context learning.

\subsection{Limitations of Tests}
It is intuitively appropriate to use the model's compositional generalization performance to reflect the model's compositionality. However, compositional generalization tests have the following limitations, which can no longer be ignored with the development of LLMs:

\textbf{(L1) Compositional generalization tests only focus on the output results. It is difficult to obtain LLMs' understanding of sample compositionality from these tests, leading to deficiencies in explainability.} 
In compositional generalization tests, we only observe the mapping results output by the model. However, there is a gap between the mapping results and the model's understanding of sample compositionality (i.e., the mapping rules that satisfy compositionality discovered by the model from the samples). This makes it difficult to obtain the model's understanding of sample compositionality from compositional generalization tests, which hinders explainable research on the compositionality of LLMs. In the pre-LLM era, model performance limitations necessitate fine-tuning, and the method of simply observing the results is widely accepted. In the era of LLMs, models have gradually evolved to be able to output contents as an externalization of their understanding in the in-context learning scenario. This development provides the possibility of obtaining the model's understanding of sample compositionality and makes the need to solve this limitation of compositional generalization tests more urgent.

\textbf{(L2) Compositional generalization tests rely on dataset partition to form the test set with combinations unseen in the training set. However, combination leakage issues can cause the requirement for unseen combinations in the test set to fail.}
The key to compositional generalization tests lies in the partition of the training set and the test set, so that the combinations in the test set have not appeared in the training set. Before the emergence of pre-trained models, for a model trained only on the partitioned training set, the combinations unseen during training can be considered truly unseen by the model, so it is appropriate to measure compositionality using the performance on the test set. However, after the emergence of pre-trained models, the validity of this evaluation has been affected. A pre-trained model may have seen combinations in the test set during the pre-training stage (i.e., combination leakage), which causes the requirement for unseen combinations in the test to fail. With the development of LLMs, the amount of pre-trained corpus has gradually increased, and the number of element combinations that LLMs have seen has also increased, making the impact of combination leakage on compositional generalization tests impossible to ignore.

\newcommand{\p}{$\mathrm{P}$\xspace}
\newcommand{\pp}{$\mathrm{P}^+$\xspace}

\begin{figure*}[h]
    \centering
    \includegraphics[width=1.0\textwidth]{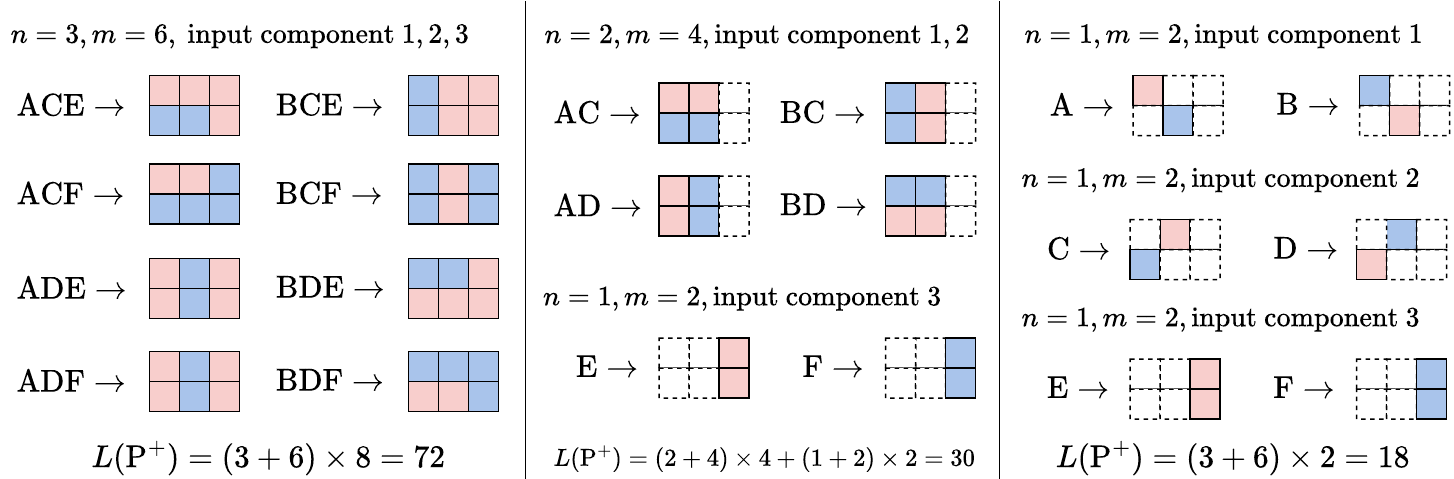}
    \caption{Examples of the mapping table of \pp ($N=3$, $M=6$, $U=2$, $d=8$). We group mappings involving the same atomic input components and mark the involved atomic output components with colors. The leftmost and rightmost examples demonstrate zero and sufficient compositionality.}
    \label{fig:map_tab}
\end{figure*}

\section{Rule-Generation Perspective}

In this section, we propose the rule-generation perspective for  compositionality estimation for LLMs. We introduce the rationale and formulation of the perspective, and the characterization of the compositionality of LLMs provided by quantitative metrics. We show that this perspective addresses the limitations of compositional generalization tests.

\subsection{Rationale}

For the limitations of compositional generalization tests, (1) the key to addressing \textbf{L1} is to obtain LLMs' understanding of sample compositionality and to quantify the compositionality exhibited by this understanding using appropriate methods, and (2) the key to addressing \textbf{L2} is to move away from the partition-based paradigm and abandon the concept of combination visibility.

To address the limitations of compositional generalization tests, we shift the perspective from result generation to rule generation, requiring LLMs to output rules that generate a set of samples (as in Figure~\ref{fig:illustration}). We want the explanation to be presented in a formal language with unambiguity, and the LLMs need no additional guidance for the generation of this formal language. Therefore, we choose a common programming language as the formal language of the explanation and ask the LLMs to output the program as the explanation. Specifically, for the set $D = \{(x_i, y_i) \ | \ x_i \in X,y_i=f(x_i)\}_{i=1}^{d}$ generated by a compositional function $f$ containing $d$ samples that cover all possible input component values, we ask the LLMs to output a program \p satisfying that for any $i \in \{1,...,d\}$, the program \p outputs $y_i$ on input $x_i$.

To estimate the compositionality of LLMs via the program \p, we introduce the complexity-based theory of compositionality proposed by \citet{complexity-based}. The theory is based on Kolmogorov Complexity $\mathcal{K}$ \cite{kolmogorov} for a quantitative definition of the compositionality of mappings from a compression perspective. For object lists $I$ and $O$, $\mathcal{K}(O)$ denotes the length of the shortest program (in a certain programming language) that outputs $O$, and $\mathcal{K}(O|I)$ denotes the length of the shortest program that outputs $O$ with input $I$. Let $D_X=\{x_i\}_{i=1}^{d}$ and $D_Y=\{y_i\}_{i=1}^{d}$ be lists of $x_i$ and $y_i$ in $D$, respectively. In this theory, the compositionality of the set $D$ (regarded as a mapping from $D_X$ to $D_Y$) is defined as $\frac{\mathcal{K}(D_Y)}{\mathcal{K}(D_Y|D_X)}$, which intuitively means the extent to which the representation of $D_Y$ can be compressed using $D_X$.

Although $\mathcal{K}$ is not computable, its upper bound can be estimated. The compositionality of an LLM can be characterized as how small an estimate of the upper bound on $\mathcal{K}(D_Y|D_X)$ is provided by the program \p that the LLM generates, as smaller estimates indicate a stronger degree of compression. The most direct upper bound estimate provided by a correct \p is the length itself. However, the length of \p is affected by many non-essential factors (e.g., formatting, naming, different description of the same process, etc.), and \p may be incorrect on $D$ (i.e., for some input $x_i$, the output is not $y_i$), so the upper bounds provided by different \p with their lengths may lack comparability. We can transform \p into a hypothetical program \pp in a uniform programming paradigm such that \pp is correct on $D$ and the upper bound estimates provided by different \pp are comparable. The upper bound estimates provided by \pp can then be used as a basis for estimating the compositionality of LLMs.

\subsection{Formulation}

Suppose a sample contains $N$ atomic input components and $M$ atomic output components. A hypothetical program \pp contains a mapping table consisting of $z$ mappings. The $z$-th mapping maps the values of $n_z$ input components to the values of $m_z$ atomic output components. Using the mapping table, \pp transforms the input into output components and combines them into an output by a fixed algorithm. Assuming that the values of all atomic input and output components are programmed with length $1$, we have that the length of \pp is $w_1\cdot\sum_{z=1}^{Z}(n_z+m_z) + w_2$, where $w_1,w_2$ are constants that are consistent for any \pp. Thus we define the size of the mapping table as a comparable metric for the estimates provided by \pp:
\begin{equation}
L(\text{\pp})=\sum_{z=1}^{Z}(n_z+ m_z)
\end{equation}

Figure~\ref{fig:map_tab} illustrates the meaning of $L(\text{\pp})$. By parsing \p for locations involving values of the input and output components (see Appendix~\ref{sec:parsing} for details), we can get the metric $L(\text{\pp})$ of its corresponding \pp. We also need to check the correctness of \p on $D$. If there are $E(\text{\p})$ errors (i.e., $E(\text{\p})=\sum_{i=1}^{d}[\text{\p}(x_i)\not=y_i]$), then $E(\text{\p})$ mappings directly corresponding to the error samples (all $N$ atomic input components mapped to all $M$ atomic output components) will need to be added to the mapping table to make \pp correct, and so $L(\text{\pp})$ increases by $(N+M)\cdot E(\text{\p})$. We thus obtain a metric for the estimates provided by \p:
\begin{equation}
L(\text{\p}) = L(\text{\pp})+(N+M)\cdot E(\text{\p})
\end{equation}

For a \pp that is correct on $D$, there are two bounds on the size of its mapping table: (1) each atomic input component corresponds to a fixed set of atomic output components (mutually disjoint and the union is all atomic output components), each mapping is a mapping of the value of an atomic input component to the value of its corresponding atomic output components, and the mapping table size $L_{\text{s}}=U(N+M)$ corresponds to sufficient compositionality; (2) the mapping table contains $d$ mappings directly corresponding to $d$ samples, and the mapping table size $L_{\text{z}}=d(N+M)$ corresponds to zero compositionality. Figure~\ref{fig:map_tab} illustrates examples of the two bounds. If $d>U$ (i.e., at least one of the values appears more than once in $D$), we can normalize the metric:
\begin{equation}
\mathcal{C}(\text{\p})=100 \cdot\frac{L_{\text{z}}-\text{Clip}(L(\text{\p}), L_{\text{z}}, L_{\text{s}})}{L_{\text{z}}-L_{\text{s}}}
\end{equation}
where Clip takes the value $L(\text{\p})$ when $L(\text{\p})\in[L_\text{z},L_\text{s}]$, and otherwise is the one closer to $L(\text{\p})$ among $L_\text{z}$ and $L_\text{s}$. We have $\mathcal{C}(\text{\p}) \in [0, 100]$. A larger metric $\mathcal{C}(\text{\p})$ represents a smaller estimate of $\mathcal{K}(D_Y|D_X)$ provided by \p, reflecting a stronger compositionality of the LLM.

\subsection{Characterization}

In the rule-generation perspective, $L(\text{\pp})$ is consistent with our intuition about compositionality, as smaller $\sum{n_z}$ indicates that the LLM is more aware of the independence of the input components, and smaller $\sum{m_z}$ indicates that the LLM more accurately identifies the output components that are influenced by the input components. Also, the characterization of the degree of compression by $L(\text{\pp})$ is consistent with our intuition about generalization. $L(\text{\p})$ further takes the number of errors $E(\text{\p})$ into account and is normalized to the metric $\mathcal{C}(\text{\p})$.
The combination of $L(\text{\pp})$ and $E(\text{\p})$ can provide a holistic and relative characterization of the compositionality of LLMs into types:

\textbf{(T1) Low $L(\text{\pp})$ and Low $E(\text{\p})$}. LLMs exhibit sufficient compositionality: they adequately and correctly capture the compositionality of the samples and utilize it to describe the rules for $D$.

\textbf{(T2) High $L(\text{\pp})$ and Low $E(\text{\p})$}. LLMs do not adequately capture the compositionality of the samples and therefore choose to low-compressively but high-correctly describe the rules for $D$ (e.g., simply using all samples directly in the program).

\textbf{(T3) Low $L(\text{\pp})$ and High $E(\text{\p})$}. LLMs do not adequately capture the compositionality of the sample, but still try to highly compress the rules of $D$, leading to a highly erroneous description.

Under this characterization, $L(\text{\pp})$ and $E(\text{\p})$ are two dimensions that characterize the degree of compression and compression loss, respectively. The high $\mathcal{C}(\text{\p})$ exhibited by \textbf{T1} can be thought of as a low-loss high compression of samples through their compositionality. The low $\mathcal{C}(\text{\p})$ exhibited by \textbf{T2} and \textbf{T3} both manifestations of the inability to correctly capture the compositionality of the samples, but they exhibit different biases: \textbf{T2} is biased towards low loss and \textbf{T3} is biased towards high compression. %

\subsection{Addressing Limitations}

The rule-generation perspective addresses \textbf{L1} and \textbf{L2} of compositional generalization tests:

(1) In the rule-generation perspective, the rules generated by LLMs directly reflect their understanding of sample compositionality. This understanding acquisition allows for more explainable estimation and analysis, thereby addressing \textbf{L1} to some extent.

(2) The rule-generation perspective is partition-free. In this perspective, the dataset is fully exposed to LLMs for rule generation, with no concept of training-test partition nor combination visibility. This partition-free paradigm mitigates the impact of combination leakage on the estimation, thereby addressing \textbf{L2} to some extent.

\begin{figure}[t]
    \centering
    \includegraphics[width=\columnwidth]{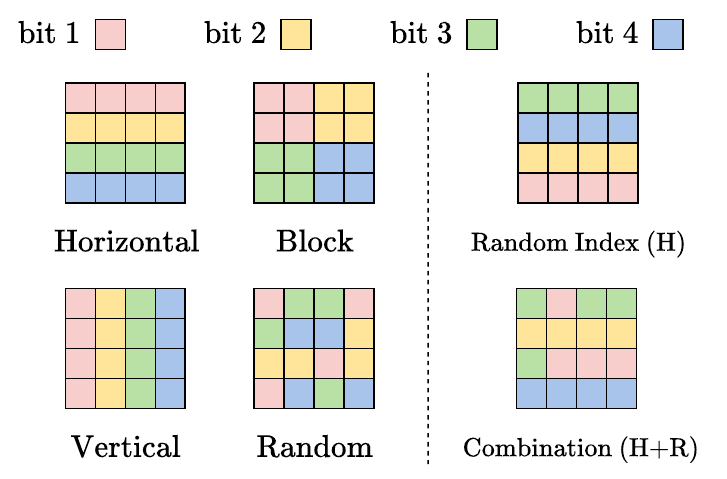}
    \caption{An illustration of the experimental settings. The color of each grid point indicates the input string bit that determines its value.}
    \label{fig:setting}
\end{figure}

\begin{table*}[]
\centering
\resizebox{\textwidth}{!}
{
\begin{tabular}{l|ccc|ccc|ccc|ccc}
\hline
            & \multicolumn{3}{c|}{\textbf{Horizontal}}                   & \multicolumn{3}{c|}{\textbf{Block}}                        & \multicolumn{3}{c|}{\textbf{Vertical}}                     & \multicolumn{3}{c}{\textbf{Random}}                        \\
            & \ $L(\text{\pp})$ & $E(\text{\p})$           & \ $\mathcal{C}(\text{\p})$ & $L(\text{\pp})$ & $E(\text{\p})$           & \ $\mathcal{C}(\text{\p})$ & $L(\text{\pp})$ & $E(\text{\p})$           & \ $\mathcal{C}(\text{\p})$ & $L(\text{\pp})$ & $E(\text{\p})$           & \ $\mathcal{C}(\text{\p})$ \\ \hline
DeepSeek-R1 & 43.23           & 1.33          & 89.80                    & 254.43          & 3.63          & 7.62                     & 254.77          & 6.00          & 0.00                     & 270.00          & 5.13          & 0.00                     \\
QwQ-Plus    & 71.33           & 7.30          & 44.00                    & 118.33          & 11.23         & 2.86                     & \textbf{95.97}  & 9.17          & 23.31                    & \textbf{181.27} & 9.20          & 0.00                     \\
o1-mini     & 49.43           & 0.30          & 94.49                    & 215.23          & 3.03          & 19.38                    & 280.87          & 2.23          & 4.29                     & 278.53          & 3.20          & 0.00                     \\
o3-mini     & 46.73           & \textbf{0.27} & \textbf{95.69}           & 150.87          & 0.47          & \textbf{57.07}           & 243.53          & 0.07          & 27.31                    & 318.13          & \textbf{0.07} & 0.67                     \\
Gemini-2.5  & 45.83           & 0.70          & 92.92                    & 197.03          & \textbf{0.13} & 42.96                    & 234.90          & \textbf{0.00} & \textbf{30.39}           & 292.00          & 0.10          & \textbf{10.00}           \\
Claude-3.7  & \textbf{40.27}  & 2.20          & 84.67                    & \textbf{89.17}  & 6.40          & 47.57                    & 216.30          & 5.10          & 14.71                    & 192.33          & 8.63          & 3.52                     \\ \hline
DeepSeek-V3 & 165.27          & \textbf{3.20} & 42.87                    & 241.53          & 6.60          & 0.00                     & 258.87          & 5.13          & 0.00                     & 239.70          & 5.90          & 3.33                     \\
Qwen-Max    & \textbf{44.57}  & 8.53          & \textbf{46.67}           & \textbf{50.70}  & 15.03         & \textbf{0.48}            & \textbf{71.80}  & 14.93         & 0.00                     & \textbf{62.00}  & 14.77         & 0.00                     \\
GPT-4o      & 46.67           & 15.70         & 0.00                     & 51.80           & 15.90         & 0.00                     & 90.63           & 15.90         & 0.00                     & 111.33          & 15.93         & 0.00                     \\
Gemini-2.0  & 189.60          & 7.17          & 7.77                     & 255.93          & \textbf{4.50} & 0.43                     & 286.13          & \textbf{2.13} & \textbf{0.67}            & 295.00          & \textbf{1.17} & \textbf{3.71}            \\
Claude-3.5  & 118.43          & 14.10         & 6.69                     & 135.03          & 15.10         & 0.00                     & 120.00          & 16.00         & 0.00                     & 133.97          & 15.43         & 0.00                     \\ \hline

\end{tabular}
}
\caption{\label{tab:base_expr}Results of the base experiment.}
\end{table*}

\section{Experiments and Analysis}
\label{sec:expr}

\subsection{Experimental settings}

\noindent \textbf{Task Formulation.} The input of the compositional function is a string of length $N=4$ and the output is a $4\times 4$ grid $(M=16)$. Each grid point has $2$ possible values $[\cdot,*]$. Each bit of the string has $U=2$ possible values, and the possible values of the $i$-th bit are the $(2i-1)$-th and $(2i)$-th uppercase letters. Each bit of the string determines the value of $4$ grid points, and the set of grid points determined by each bit of the string is mutually exclusive. The value of any grid point differs when the value of the input bit that determines it differs. All possible $d=16$ samples generated by the compositional function are provided to the LLMs as $D$ and the LLMs are asked to generate a Python program to describe the generation rules for $D$.

\noindent \textbf{Data Generation.} Our base experiment consists of four different compositional function settings: (1) Horizontal (the $i$-th bit determines the $i$-th row), (2) Block (the $i$-th bit determines the $i$-th $2\times2$ subgrid), (3) Vertical (the $i$-th bit determines the $i$-th column), and (4) Random (each bit determines $4$ random grid points).  For each setting, we sample $30$ different compositional functions for data generation and report the average results of LLMs on the task. Our extended experimental settings, including random index and setting combination, are discussed further in~\ref{sec:extend}. Figure~\ref{fig:setting} shows an illustration of the settings.

\noindent \textbf{Evaluated LLMs.} The LLMs we evaluate include the reasoning models: DeepSeek-R1-0120 \cite{deepseekai2025deepseekr1incentivizingreasoningcapability}, QwQ-Plus-0305 \cite{qwq32b}, o1-mini-2024-09-12 \cite{openai2024openaio1card}, o3-mini-2025-01-31 \cite{o3-mini}, Gemini-2.5-pro-exp-03-25 \cite{gemini-2.5}, Claude-3.7-Sonnet-20250219 \cite{claude-3.7}, and the non-reasoning models: DeepSeek-V3-0324 \cite{deepseekai2025deepseekv3technicalreport}, Qwen-Max-0125 \cite{qwen2.5}, GPT-4o-2024-08-06 \cite{openai2024gpt4ocard}, Gemini-2.0-flash \cite{gemini-2.0}, and Claude-3.5-Haiku-20241022 \cite{claude-3.5}.

\begin{table*}[]
\centering
\resizebox{\textwidth}{!}
{
\begin{tabular}{l|ccc|ccc}
\hline
            & \multicolumn{3}{c|}{\textbf{Random Index (H)}}                                & \multicolumn{3}{c}{\textbf{Setting Combination (H+R)}}                           \\
            & $L(\text{\pp})$        & $E(\text{\p})$                   & $\mathcal{C}(\text{\p})$ & $L(\text{\pp})$         & $E(\text{\p})$                   & $\mathcal{C}(\text{\p})$ \\ \hline
DeepSeek-R1 & 174.07 (+130.83)       & 4.17 (+2.83)          & 26.54 (-63.26)           & 187.30 (+30.68)         & 3.57 (+0.33)          & 27.43 (-17.47)           \\
QwQ-Plus    & 138.23 (+66.90)        & 10.97 (+3.67)         & 0.00 (-44.00)            & 147.60 (+21.30)         & 9.83 (+1.58)          & 4.00 (-18.00)            \\
o1-mini     & 138.27 (+88.83)        & 1.27 (+0.97)          & 56.69 (-37.80)           & 176.60 (+12.62)         & 2.67 (+0.92)          & 34.26 (-12.98)           \\
o3-mini     & 106.90 (+60.17)        & \textbf{0.50} {(+0.23)} & 73.20 (-22.49)           & 84.73 (-97.70)          & 0.77 (+0.60)          & \textbf{78.79} {(+30.61)}  \\
Gemini-2.5  & 87.57 (+41.73)         & 0.93 (+0.23)          & 76.58 (-16.33)           & 205.53 (+36.62)         & \textbf{0.73} {(+0.33)} & 38.98 (-12.48)           \\
Claude-3.7  & \textbf{40.23} {(-0.03)} & 3.13 (+0.93)          & \textbf{79.04} {(-5.63)}   & \textbf{62.57} {(-53.73)} & 3.80 (-1.62)          & 66.39 (+22.30)           \\ \hline
DeepSeek-V3 & 145.40 (-19.87)        & 8.57 (+5.37)          & \textbf{12.85} {(-30.02)}  & 207.43 (+4.95)          & 5.27 (+0.72)          & \textbf{9.89} {(-13.21)}   \\
Qwen-Max    & 51.00 (+6.43)          & 14.83 (+6.30)         & 1.38 (-45.29)            & 58.93 {(+5.65)}  & 13.67 (+2.02)         & 6.67 (-16.67)            \\
GPT-4o      & \textbf{45.33} {(-1.33)} & 15.80 (+0.10)         & 0.00 (-0.00)             & \textbf{58.20} (-20.80)          & 15.80 (-0.02)         & 0.00 (-0.00)             \\
Gemini-2.0  & 267.10 (+77.50)        & \textbf{3.83} {(-3.33)} & 1.81 (-5.96)             & 267.53 (+25.23)         & \textbf{4.07} {(-0.10)} & 0.43 (-5.32)             \\
Claude-3.5  & 116.37 (-2.07)         & 15.73 (+1.63)         & 0.00 (-6.69)             & 113.60 (-12.60)         & 15.83 (+1.07)         & 0.00 (-3.35)             \\ \hline

\end{tabular}
}
\caption{\label{tab:extend}Results of extended experiments. The amount of change compared to the results of the base experiment is shown in parentheses (left: compared to Horizontal; right: compared to the average of Horizontal and Random).}
\end{table*}

\subsection{Base Experiment}
Table~\ref{tab:base_expr} shows the results of the base experiment.

\subsubsection{Compositionality Characterization}
The non-reasoning models we evaluate exhibit fairly low compositionality in most cases, and only DeepSeek-V3 and Qwen-Max exhibit relatively high compositionality in the Horizontal setting. Relatively among the non-reasoning models: (1) Qwen-Max, GPT-4o, and Claude-3.5 are characterized as \textbf{T3}. Although they exhibit strong compression, their extremely high error rate means that their descriptions of the rules for $D$ are almost completely incorrect. (2) Deepseek-V3 and Gemini-2.0 are characterized as \textbf{T2}. They have a relatively high degree of correctness in describing the rules for $D$, but also a relatively low degree of compression.

The reasoning models we evaluate generally exhibit stronger compositionality than non-reasoning models in settings other than Random. The stronger compositionality stems from maintaining a certain degree of compression at a generally lower number of errors. However, the reasoning models do not exhibit sufficient compositionality in settings other than Horizontal. In settings other than Horizontal, the reasoning models exhibit the following characterization in relative terms: (1) QwQ-Plus and Claude-3.7 are characterized as \textbf{T3}. They exhibit a high error rate and a high degree of compression. (2) Gemini-2.5 and o3-mini are characterized as \textbf{T2}. They exhibit a low error rate and a low degree of compression. %
(3) DeepSeek-R1 and o1-mini are characterized roughly between \textbf{T2} and \textbf{T3}.

\subsubsection{Impact of the Settings}
\label{sec:impact_setting}
Since all possible samples are provided in $D$, for any bit of the input string, LLMs can theoretically find that the bit independently determines some grid points from multiple pairs of samples differing only on that bit, which is independent of the compositional function setting. However, the $\mathcal{C}(\text{\p})$ of LLMs differ clearly across settings and mostly follow the relation: Horizontal $>$ Block $>$ Vertical $>$ Random (except for QwQ-Plus which shows Vertical $>$ Block).

We hypothesize that the compositionality exhibited by LLMs on this task is influenced by how intuitive the sample's compositionality is. Of the four compositional function settings, the first three have a certain regularity and a similar intuition in the two-dimensional view, since the grid points determined by each bit of the string in these settings are connected in the two-dimensional plane. However, in the linear form of text input, for the continuity of the grid point positions determined by each bit of the string in the settings, we have Horizontal $>$ Block $>$ Vertical. As continuity declines, the intuition of the compositionality of samples may decline for LLMs, which are used to intuitively capturing by row. The Random setting, on the other hand, provides no intuition for LLMs at all.

\subsection{Extended Experiments}
\label{sec:extend}
Table~\ref{tab:extend} shows the results of the extended experiments.

\subsubsection{Random Index}
We conduct extended experiments with the Random Index setting on the Horizontal setting, where LLMs exhibit the strongest compositionality in the base experiment. In the base experiment, the $i$-th bit of the input string corresponds to the $i$-th row of the grid in the Horizontal setting. With the extended Random Index setting, the row corresponding to each bit is randomized.

The results show that the Random Index setting causes a severe weakening of the compositionality exhibited by the LLMs. For reasoning models, all models except Claude-3.7 exhibit high $L(\text{\pp})$ increases, indicating a reduction in compression. For reasoning models, all models except Claude-3.7 exhibit a high increase in $L(\text{\pp})$, indicating a decrease in compression; all models exhibit an increase in $E(\text{\p})$, indicating an elevated compression loss, especially DeepSeek-R1 and QwQ-Plus. Among the reasoning models, Claude-3.7 exhibits the least decrease in compositionality and shows the strongest compositionality with the Random Index setting. Most of the non-reasoning models show a decrease in compositionality, approaching zero compositionality.

Although Random Index does not change the Horizontal pattern followed by each component mapping, the LLMs generally show a decline in compositionality, which is partly indicative of the LLMs' reliance on sequential correspondences for sample compositionality capture for this task, reflecting a deficiency in compositionality.

\subsubsection{Setting Combination}
We combine Horizontal and Random, the two settings where LLMs exhibit the strongest and weakest compositionality in the base experiment. Under the setting combination, two random rows in the grid use the Horizontal setting, and the other grid points use the Random setting.

For the reasoning models, compared to the average of the metrics in the two settings: (1) DeepSeek-R1, QwQ-Plus, o1-mini, and Gemini-2.5 exhibit elevated $L(\text{\pp})$ and $E(\text{\p})$ and decreased $\mathcal{C}(\text{\p})$. This means that when the compositionality of a portion of the sample's components (Random) is difficult to capture, their degree of compression and compression loss for all components are affected, even though the compositionality of the remaining components (Horizontal) is relatively easy for them to capture. This reflects a compositionality flaw of LLMs in that they have difficulty in independent compositionality capture for different sets of components. (2) Claude-3.7 and o3-mini exhibit elevated $\mathcal{C}(\text{\p})$, which suggests that they are somewhat capable of independent compositionality capture for the Horizontal component. In this case, even if they still exhibit low compression in the Random portion, there is a clear $L(\text{\pp})$ reduction brought about by the reduction of the component space. In addition, Claude-3.7 also exhibits a decrease in $E(\text{\p})$, which indicates that its compression loss can also be reduced as the component space is reduced. The non-reasoning models mostly exhibit a decrease in $\mathcal{C}(\text{\p})$, approaching zero compositionality. Overall, many of the LLMs exhibit deficiencies in independent compositionality capture for different sets of components.

\subsection{Statistical Analysis}
In Appendix~\ref{app:stat}, we supplement the statistical analysis of the base experiments and extended experiments, including the standard deviation of the results and significance tests for cross-model performance comparisons.

\subsection{Qualitative Analysis}
\label{sec:qua}
Figure~\ref{fig:code} (in Appendix~\ref{app:example}) shows fragments of some of the programs generated by LLMs.

(1) There are some examples of high $\mathcal{C}(\text{\p})$ in settings other than Random. The output strings determined by a bit of the input string typically each correspond to at most one segment of a contiguous region within one row of the grid in linear form. This partly supports our hypothesis in~\ref{sec:impact_setting}.

(2) Typical examples of high $L(\text{\pp})$ and low $E(\text{\p})$ are simply enumerating all samples in all $D$. Typical examples of low $L(\text{\pp})$ and high $E(\text{\p})$ are compression using simple algorithms not fully supported by $D$.

(3) High $\mathcal{C}(\text{\p})$ with Random Index setting is exemplified by the perception of sequential non-correspondence, and high $\mathcal{C}(\text{\p})$ with Setting Combination is exemplified by the perception of regions independently affected by different settings (see Appendix~\ref{app:example} for details). In the extended experiments, typical examples of high $E(\text{\p})$ are still generating programs according to the Horizontal setting in the base experiment; typical examples of high $L(\text{\pp})$ are the same as in (2), arising from the inability to capture compositionality due to the extended settings.

\section{Discussion on Controversies}

In this section, we discuss some of the controversies surrounding the rule-generation perspective and present the relevant experimental results. Due to limitations in the availability of the official platform API, our experiments on DeepSeek are instead modified to evaluate both the reasoning and non-reasoning versions of DeepSeek-V3.2 (denoted as DeepSeek-V3.2$_\text{R}$ and DeepSeek-V3.2$_\text{R}$, respectively; experiments in Section~\ref{sec:expr} are re-run), which together with the Qwen series models constitute the results.

\label{sec:discuss}

\begin{table*}[h]
\centering
\resizebox{\textwidth}{!}
{
\begin{tabular}{l|cccc|cccc|cccc}
\hline
            & \multicolumn{4}{c|}{\textbf{Horizontal}}                   & \multicolumn{4}{c|}{\textbf{Block}}                        & \multicolumn{4}{c}{\textbf{Vertical}}                                 \\
            & \ $L(\text{\pp})$ & $E(\text{\p})$           & \ $\mathcal{C}(\text{\p})$ & $ \mathcal{A} $ & \ $L(\text{\pp})$ & $E(\text{\p})$           & \ $\mathcal{C}(\text{\p})$ & $ \mathcal{A} $ & \ $L(\text{\pp})$ & $E(\text{\p})$           & \ $\mathcal{C}(\text{\p})$ & $ \mathcal{A} $ \\ \hline
DeepSeek-V3.2$_\text{R}$ & 81.13         & 3.47          & 64.51                    & 100.00 & 171.90 & 8.33 & 1.90 & 86.67 & 265.10 & 3.67 & 0.00 & 53.33                   \\
QwQ-Plus    & 71.33           & 7.30          & 44.00  & 70.00                  & 118.33          & 11.23         & 2.86        & 10.00             & 95.97  & 9.17          & 23.31    & 10.00 
                    \\ 
DeepSeek-V3.2$_\text{N}$ & 145.50 & 5.40 & 30.95 & 76.67 & 238.13 & 5.50 & 0.00 & 26.67 & 279.20 & 3.80 & 0.00 & 6.67         \\
Qwen-Max    & 44.57  & 8.53          & 46.67 & 100.00           & 50.70  & 15.03         & 0.48 & 53.33            & 71.80  & 14.93         & 0.00 & 0.00                             \\
\hline
            & \multicolumn{4}{c|}{\textbf{Random}}                   & \multicolumn{4}{c|}{\textbf{RI (H)}}                        & \multicolumn{4}{c}{\textbf{SC (H+R)}}                                 \\
            & \ $L(\text{\pp})$ & $E(\text{\p})$           & \ $\mathcal{C}(\text{\p})$ & $ \mathcal{A} $ & \ $L(\text{\pp})$ & $E(\text{\p})$           & \ $\mathcal{C}(\text{\p})$ & $ \mathcal{A} $ & \ $L(\text{\pp})$ & $E(\text{\p})$           & \ $\mathcal{C}(\text{\p})$ & $ \mathcal{A} $ \\ \hline
DeepSeek-V3.2$_\text{R}$ & 231.57 & 6.30 & 0.00 & 30.00 & 145.40 & 9.50 & 13.23 & 80.00 & 217.90 & 5.70 & 6.14 & 63.33      \\
QwQ-Plus    & 181.27 & 9.20          & 0.00   & 0.00    & 138.23        & 10.97         & 0.00  & 20.00          & 147.60         & 9.83         & 4.00    &  20.00
                    \\ 
DeepSeek-V3.2$_\text{N}$  & 251.23 & 5.37 & 0.00 & 10.00 & 234.50 &  7.53 & 0.10 & 40.00 & 240.10 & 5.50 & 1.20 & 23.33     \\
Qwen-Max    & 62.00  & 14.77         & 0.00  & 0.00 & 51.00          & 14.83        & 1.38 & 63.33          & 58.93  & 13.67         & 6.67  & 50.00           \\
\hline

\end{tabular}
}
\caption{\label{tab:cg_comp}Comparison of the result-generation perspective and the rule-generation perspective. $\mathcal{A}$ denotes accuracy under the result-generation perspective. \textbf{RI} refers to the random index. \textbf{SC} refers to the setting combination.}
\end{table*}

\subsection{Programming or Natural Language?}
In this work, we use common programming languages as the language for describing rules in order to achieve unambiguous and automatic estimation. However, this raises two potential issues: 

(1) The programming capabilities of the model can potentially influence the results. To avoid this issue, we are currently limited to selecting a simple string-to-grid task for experiments, where the model can correctly translate rules into programs. To validate the impact of LLMs' programming capabilities on estimation in the string-to-grid task, we evaluate their ability to process the task when rules are provided directly. Specifically, we use natural language to fully describe the rules generating $D$, then instruct LLMs to describe these rules using programs. We evaluate LLMs under the Random setting where they exhibit weak compositionality in the program-generation perspective, and find that LLMs perform well on the task. The reasoning models DeepSeek-V3.2$_\text{R}$ and QwQ-Plus both achieve 100\% accuracy; the non-reasoning models DeepSeek-V3.2$_\text{N}$ and Qwen-Max both achieve 98.33\% accuracy. Based on the experimental results, we believe the model possesses sufficient programming capabilities for the task and does not adversely affect compositionality estimation.

(2) The conversion from program to mapping table is heuristic (as in Appendix~\ref{sec:parsing}). It is difficult for heuristic rules to cover all cases completely, despite our efforts to cover all cases we have encountered. %
When using heuristic rules, in addition to the potential need for new heuristic rules for new models and tasks, the sensitivity of LLMs to prompts may also affect compositionality estimateion, as new programs triggered by new prompts may lead to cases not covered by existing heuristic rules. We conduct prompt sensitivity tests under our current experimental settings and find that the sensitivity of LLMs to prompts does not have a significant impact on the current experimental conclusions, thereby temporarily ruling out this issue (see Appendix~\ref{app:prompt} for details).

Compared to using programming languages, using natural language as a rule description language avoids controversies arising from programming capabilities. However, quantifying compositionality under natural language with ambiguity becomes more challenging, potentially requiring the introduction of more costly manual evaluations. How to more generally quantify compositionality via programs, and how to quantify compositionality via natural language descriptions, are future research directions that remain to be explored.

\subsection{Result Analysis or Rule Analysis?}
The potential differences in the internal mechanisms of LLMs for result generation and rule generation can lead to different compositionality estimations. We present LLMs with 8 samples from dataset $D$ and ask them to generate outputs for unseen inputs, thereby evaluating their accuracy from a result-generation perspective. The results and comparison with the rule-generation perspective are shown in Table~\ref{tab:cg_comp}.

The estimation from the result-generation perspective is typically higher than that from the rule-generation perspective. For example, under the Horizontal setting, some models can perfectly achieve compositional generalization in result generation, yet fail to demonstrate high compositionality under the rule-generation perspective. This generally aligns with our intuition that rule generation is more challenging than result generation.

The estimations from the result-generation perspective and the rule-generation perspective are not significantly correlated. Under a given setting, an LLM with a higher $\mathcal{A}$ does not necessarily exhibit a higher $\mathcal{C}(\text{\p})$. %
Since there is no evidence demonstrating a correlation between the rules generated under the rule-generation perspective and the process of generating results under the result-generation perspective, these results are unsurprising. Further exploration of the differences in the internal mechanisms of LLMs regarding result generation and rule generation is needed to explain the inconsistencies in the results.

Researchers have not reached a consensus on whether compositionality research requires only result analysis or the introduction of rule analysis \citep{compositionality-survey}. Our work cannot lead to consensus, but it introduces a novel perspective for future exploration. Given the advantages of explainability and partition-free nature, we believe continued exploration of the rule-generation perspective holds significant potential in this field.

\section{Conclusion}
In this work, we propose the rule-generation perspective for compositionality estimation for LLMs. This more explainable and partition-free perspective addresses the limitations of compositional generalization tests and provides a new way to analyze the compositionality characterization of LLMs. Through experiments and analysis based on this perspective, we identify different compositionality characterizations and compositionality defects exhibited by existing advanced LLMs. This perspective provides support for the study of compositionality of LLMs.

\section*{Limitations}

As discussed in Section~\ref{sec:discuss}, we currently employ programming languages as rule description languages in the rule-generation perspective. However, the impact of LLMs' programming capabilities and the reliance on heuristic rules for program-to-mapping table conversion limit the practical application of this perspective to more complex tasks. As LLMs' programming capabilities advance further, experiments on more complex tasks may become feasible in this perspective. Nevertheless, the reliance on heuristic rules remains a persistent challenge. If programming languages continue to serve as the rule description language, deeper exploration into achieving automatic estimation for any task may be necessary. On the other hand, we currently lack a clear approach for appropriately estimating compositionality when using natural language as the rule description language. Compared to formal languages like programming languages, the ambiguity inherent in natural language makes compositionality estimation significantly more challenging, particularly for automatic estimation. Even when accepting more costly manual evaluations, there remains a lack of theoretical research on quantifying rules described in natural language into compositionality estimation. Relevant methods require further investigation.

As discussed in Section~\ref{sec:discuss}, a controversy exists in compositionality research regarding result and rule: does result analysis suffice, or is explainable rule analysis necessary? For LLMs, the distinction between result generation and rule generation at the internal mechanism level remains unclear, making it difficult to draw definitive conclusions about the necessity of introducing rules at this level. Furthermore, rule generation itself may not necessarily be treated as a processing mechanism for result generation, as the two may represent fundamentally different modes of thinking, which is also reflected in human cognition (e.g., we may approach multiple-choice questions and short-answer questions in tests with entirely different processing strategies). This controversy will persist until the fundamental differences between the internal mechanisms of result generation and rule generation are fully understood. Setting aside the controversies arising from unclear internal mechanisms, the rule-generation perspective offers greater explainability at the externalization level compared to the result-generation perspective. This is manifested in the direct access to LLMs' understanding of sample compositionality.

Despite the limitations and controversies, the rule-generation perspective represents a novel exploratory perspective. Compared to the compositional generalization tests long employed in compositionality research, this perspective offers greater explainability, and its inherent partition-free nature mitigates compositional leakage to some extent. We will continue exploring this perspective to achieve more general compositionality estimation for LLMs.

\section*{Ethics Statement}
\par We comply with the license to use language models for scientific research purposes only. The datasets we construct will also be open source for scientific research purposes. The datasets we use do not contain any information that names or uniquely identifies individual people or offensive content.
\par The AI assistant we use in our work is Copilot (for simple code completion).

\section*{Acknowledgements}
This work was supported by National Natural Science Foundation of China (62576010) and the Academic Research Projects of Beijing Union University (NO.ZK10202405). The corresponding author is Houfeng Wang.

\bibliography{custom}

\clearpage
\appendix

\section{Details on obtaining $L(\text{\pp})$}
\label{sec:parsing}

\par To obtain the corresponding metric $L(\text{\pp})$ from the program \p, we need to determine the $\sum{n_z}$ and $\sum{m_z}$ mapping table by the location of the values involving the input and output components in \p. The contents of the comments are ignored.

\subsection{Determination of $\sum{n_z}$}
To determine $\sum{n_z}$, we need to determine which combinations of values of the input components used to determine the output are contained in \p. A combination consists of values of input components that are (1) in a mapping of an explicit mapping table (dictionary), (2) on the same row, and (3) on a path in a nested structured tree of conditional judgments. For cases (2) and (3), it is considered to be used to determine the output when the row or the execution statements of the conditional judgments involve the values of the output components.

With the help of Python's AST tool, we are able to get all combinations of values of input components used to determine the output. The value of an input component may be on the right of an assignment statement and then affect a wider range through the variables on the left of the assignment statement. To handle this situation, we maintain the set of values of the input components involved for each variable due to assignment and utilize them when determining the values of the input components involved in the statement. For the nested structure of conditional judgments, we construct trees and obtain all possible paths and corresponding combinations by traversing them. For an \textsl{else} statement, we match it to the corresponding \textsl{if} and \textsl{elif} statements and treat it as containing one hypothetical value for each input bit involved in the \textsl{if} and \textsl{elif} statements.

The same combination may occur several times in \p and can be generalized to the same mapping. Therefore, we count the total length of the values of the input components involved in mutually exclusive combinations as $\sum{n_z}$.

\subsection{Determination of $\sum{m_z}$}
$\sum{m_z}$ can theoretically be determined by finding the values of all output components in \p and computing the length sum. However, we find that \p sometimes expresses the determination of the output indirectly in other forms (e.g., storing the coordinates of the determined grid points), which occurs mainly in the explicit mapping table (dictionary). Therefore, we perform additional processing to count each atomic unit on the right side of the explicit mapping table as a value of an output component in $\sum{m_z}$.

\subsection{Examples}

\begin{figure*}[t]
    \centering
    \includegraphics[width=\textwidth]{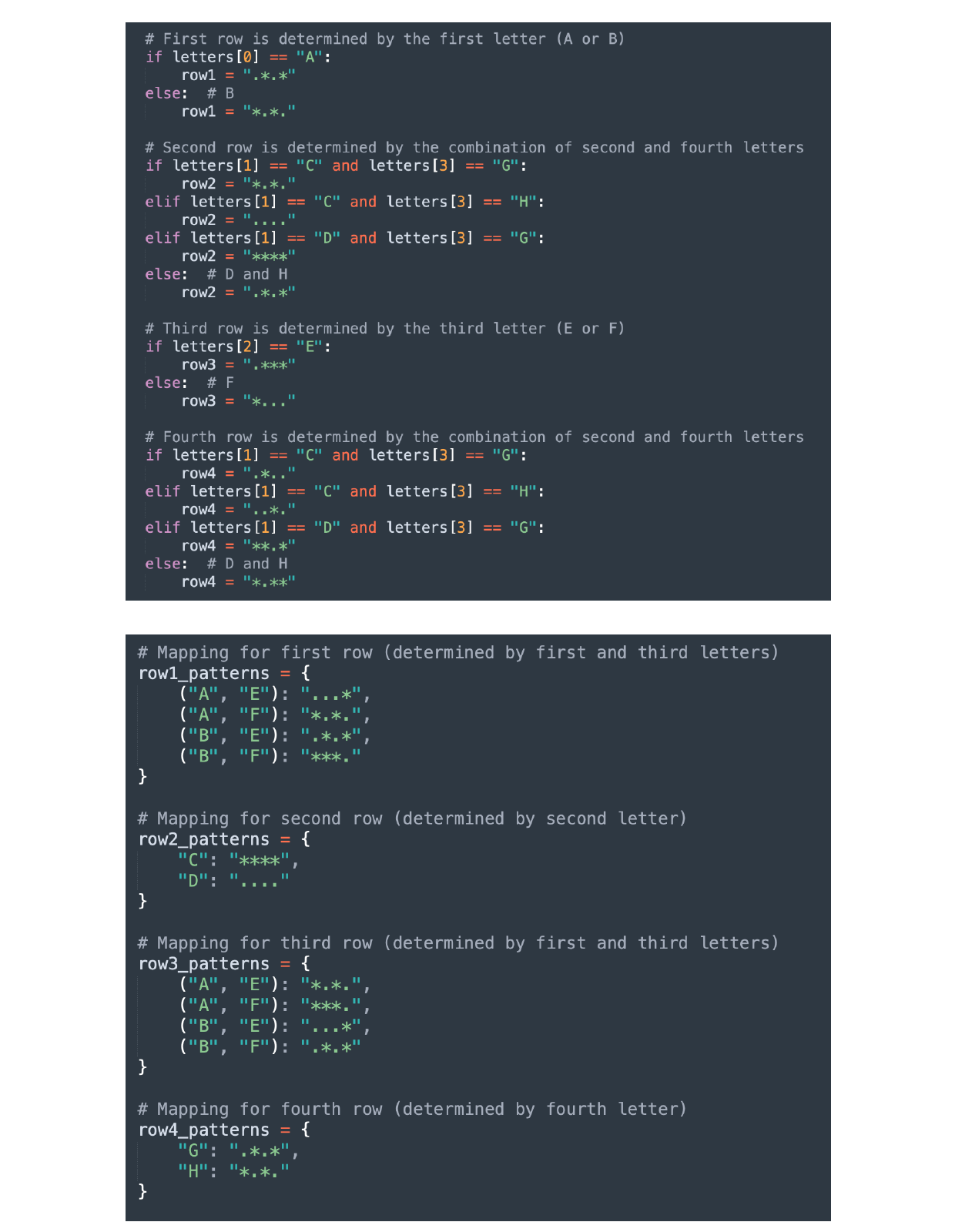}
    \caption{Two examples of fragments of programs generated by LLMs. They both have $\sum{n_z}=12$ and $\sum{m_z}=48$.}
    \label{fig:app-code}
\end{figure*}

Figure~\ref{fig:app-code} shows two example code fragments. They both have $\sum{n_z}=12$ and $\sum{m_z}=48$.

Code fragment 1 mainly shows the case with conditional judgments. The first three code blocks contribute $2,8,2$ to $\sum{n_z}$, and $8,16,8$ to $\sum{m_z}$. All combinations of input component values in the last code block have already appeared in the second code block, so they are no longer counted in $\sum{n_z}$. The last code block contributes 16 to $\sum{m_z}$.

Code fragment 2 mainly shows the case with dictionaries. The four dictionaries contribute $16,8,16,8$ to $\sum{m_z}$. The three dictionaries, except dictionary 3, contribute $8,2,2$ to $\sum{n_z}$.

\begin{figure*}[!t]
    \centering
    \includegraphics[width=\textwidth]{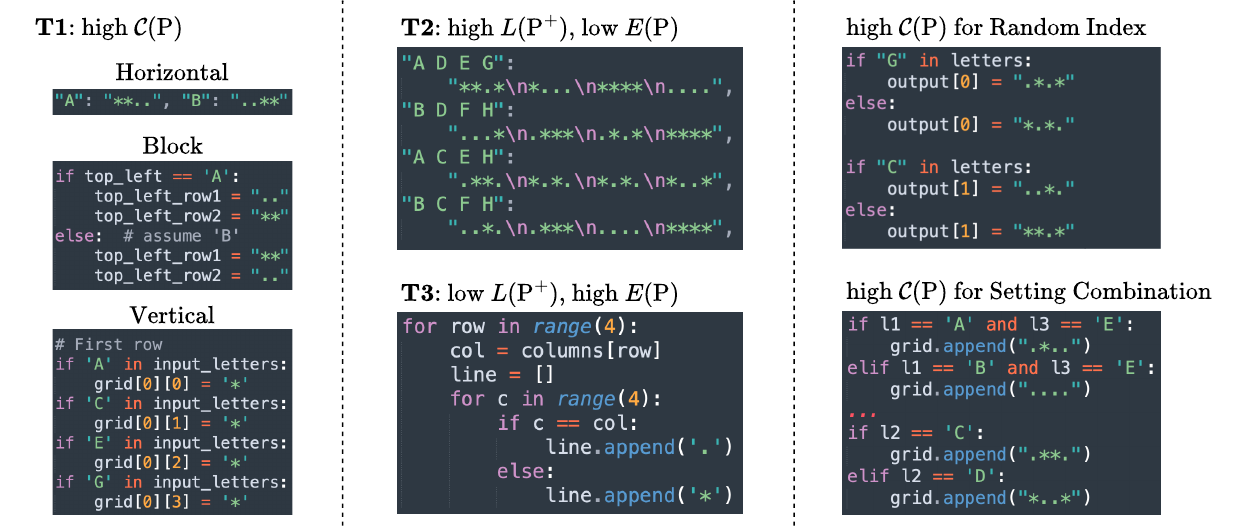}
    \caption{Examples of fragments of programs generated by LLMs.\label{fig:code}}
\end{figure*}

\section{Supplemental Results}

\subsection{Statistical Analysis}
\label{app:stat}
\begin{table}[t]
\centering
\resizebox{\columnwidth}{!}
{
\begin{tabular}{l|ccc|ccc}
\hline
                                   & \multicolumn{3}{c|}{\textbf{Reasoning}}                                                     & \multicolumn{3}{c}{\textbf{Non-Reasoning}}                                \\
\multirow{-2}{*}{\textbf{Setting}} & $L(\text{\pp})$               & $E(\text{\p})$    & $\mathcal{C}(\text{\p})$               & $L(\text{\pp})$              & $E(\text{\p})$ & $\mathcal{C}(\text{\p})$ \\ \hline
Horizontal                         & {\color[HTML]{333333} 25.06}  & {\color[HTML]{333333} 3.50} & {\color[HTML]{333333} 24.60} & {\color[HTML]{333333} 73.57} & 5.28                    & 28.52          \\
Block                              & {\color[HTML]{333333} 115.92} & {\color[HTML]{333333} 4.45} & {\color[HTML]{333333} 33.75} & {\color[HTML]{333333} 70.22} & 4.06                    & 0.99           \\
Vertical                           & {\color[HTML]{333333} 115.40} & {\color[HTML]{333333} 4.32} & {\color[HTML]{333333} 30.18} & {\color[HTML]{333333} 85.38} & 3.19                     & 0.73           \\
Random                             & {\color[HTML]{333333} 92.00}  & {\color[HTML]{333333} 4.75} & {\color[HTML]{333333} 8.53}  & {\color[HTML]{333333} 79.39} & 3.56                     & 5.19           \\
SC(H+R)               & {\color[HTML]{333333} 108.97} & {\color[HTML]{333333} 4.61} & {\color[HTML]{333333} 33.97} & {\color[HTML]{333333} 73.28} & 3.48                     & 7.43           \\
RI(H)                     & {\color[HTML]{333333} 103.63} & {\color[HTML]{333333} 4.63} & {\color[HTML]{333333} 35.34} & {\color[HTML]{333333} 70.05} & 3.51                     & 8.34           \\ \hline
\end{tabular}
}
\caption{\label{tab:devi}The average of standard deviations of $L(\text{\pp})$, $E(\text{\p})$, and $\mathcal{C}(\text{\p})$ for reasoning models and non-reasoning models under different settings.}
\end{table}

\begin{table*}[t]
\centering
\resizebox{\textwidth}{!}{
\begin{tabular}{l|l}
\hline
\textbf{Setting} & $L(\text{\pp})$ \textbf{Groups} \\
\hline
Horizontal                         &     (40.27, 43.23, 44.57, ) (45.83, 46.67, 46.73, 49.43, ) (71.33, ) (118.43, 165.27, 189.60, )  \\
Block                              &   (50.70, ) (51.80, ) (89.17, 118.33, ) (135.03, 150.87, 197.03, ) (215.23, 241.53, 254.43, 255.93, )         \\
Vertical                           &    (71.80, ) (90.63, ) (95.97, ) (120.00, ) (216.30, 234.90, 243.53, 254.77, 258.87, ) (280.87, 286.13, )       \\
Random                             &      (62.00, ) (111.33, 133.97, 181.27, 192.33, ) (239.70, 270.00, ) (278.53, 292.00, 295.00, ) (318.13, )      \\
SC(H+R)               &     (58.20, 58.93, ) (62.57, ) (84.73, ) (113.60, 147.60, 176.60, ) (187.30, 205.53, 207.43, 267.53, )       \\
RI(H)                     &    (40.23, ) (45.33, 51.00, 87.57, 106.90, ) (116.37, 138.23, 138.27, 145.40, 174.07, ) (267.10, )        \\ \hline
\textbf{Setting} & $E(\text{\p})$ \textbf{Groups} \\
\hline
Horizontal                         &     (0.27, 0.30, 0.70, 1.33, ) (2.20, 3.20, ) (7.17, 7.30, 8.53, ) (14.10, 15.70, )  \\
Block                              &   (0.13, 0.47, ) (3.03, 3.63, 4.50, 6.40, 6.60, ) (11.23, ) (15.03, 15.10, 15.90, )         \\
Vertical                           &    (0.00, 0.07, ) (2.13, 2.23, ) (5.10, 5.13, 6.00, 9.17, ) (14.93, 15.90, 16.00, )       \\
Random                             &      (0.07, 0.10, ) (1.17, 3.20, ) (5.13, 5.90, 8.63, ) (9.20, ) (14.77, 15.43, 15.93, )      \\
SC(H+R)               &     (0.73, 0.77, ) (2.67, 3.57, 3.80, 4.07, 5.27, ) (9.83, ) (13.67, ) (15.80, 15.83, )       \\
RI(H)                     &    (0.50, 0.93, 1.27, 3.13, ) (3.83, 4.17, ) (8.57, 10.97, ) (14.83, 15.73, 15.80, )        \\ \hline
\textbf{Setting} & $\mathcal{C}(\text{\p})$ \textbf{Groups} \\
\hline
Horizontal                         &     (95.69, ) (94.49, 92.92, 89.80, 84.67, ) (46.67, 44.00, 42.87, ) (7.77, 6.69, ) (0.00, )  \\
Block                              &   (57.07, 47.57, 42.96, ) (19.38, ) (7.62, 2.86, 0.48, 0.43, 0.00, 0.00, 0.00, )        \\
Vertical                           &    (30.39, 27.31, 23.31, 14.71, ) (4.29, 0.67, 0.00, 0.00, 0.00, 0.00, 0.00, )      \\
Random                             &      (10.00, 3.71, 3.52, 3.33, ) (0.67, 0.00, 0.00, 0.00, 0.00, 0.00, 0.00, )     \\
SC(H+R)               &     (78.79, 66.39, ) (38.98, 34.26, 27.43, ) (9.89, 6.67, 4.00, ) (0.43, 0.00, 0.00, )      \\
RI(H)                     &    (79.04, 76.58, 73.20, ) (56.69, ) (26.54, 12.85, ) (1.81, 1.38, 0.00, 0.00, 0.00, )        \\ \hline
\end{tabular}
}
\caption{\label{tab:group}Significance tests for cross-model performance comparisons (Mann-Whitney U test; $p<0.05$ is considered significant). We rank values from strongest to weakest for each setting and indicate the groupings (separated by brackets). Differences between results within the same group cannot be considered significant, while significant differences exist between results in different groups.}
\end{table*}

\begin{table*}[t]
\centering

\begin{tabular}{l|ccc|ccc|ccc}
\hline
                                   & \multicolumn{3}{c|}{\textbf{Test1}}                                                        & \multicolumn{3}{c|}{\textbf{Test2}}                                                  & \multicolumn{3}{c}{\textbf{Test3}}                                                   \\
 & $L(\text{\pp})$               & $E(\text{\p})$   & $\mathcal{C}(\text{\p})$     & $L(\text{\pp})$               & $E(\text{\p})$ & $\mathcal{C}(\text{\p})$ & $L(\text{\pp})$               & $E(\text{\p})$ & $\mathcal{C}(\text{\p})$ \\ \hline
DeepSeek-V3.2$_\text{R}$                  & {\color[HTML]{333333} 81.13}  & {\color[HTML]{333333} 3.47} & {\color[HTML]{333333} 64.51} & {\color[HTML]{333333} 53.30}  & 6.40                      & 57.57                    & {\color[HTML]{333333} 40.63}  & 5.86                      & 62.77                    \\
QwQ-Plus                           & {\color[HTML]{333333} 71.33}  & {\color[HTML]{333333} 7.30} & {\color[HTML]{333333} 44.00} & {\color[HTML]{333333} 45.10}  & 9.93                      & 35.39                    & {\color[HTML]{333333} 85.13}  & 6.50                      & 43.73                    \\
DeepSeek-V3.2$_\text{N}$                   & {\color[HTML]{333333} 145.50} & {\color[HTML]{333333} 5.40} & {\color[HTML]{333333} 30.95} & {\color[HTML]{333333} 120.27} & 8.77                      & 19.90                    & {\color[HTML]{333333} 155.20} & 6.00                      & 24.92                    \\
Qwen-Max                           & {\color[HTML]{333333} 44.57}  & {\color[HTML]{333333} 8.53} & {\color[HTML]{333333} 46.67} & {\color[HTML]{333333} 41.40}  & 7.93                      & 49.95                    & {\color[HTML]{333333} 40.20}  & 7.47                      & 51.51                    \\ \hline
\end{tabular}

\caption{\label{tab:sensi}Results for three different prompts under the Horizontal setting.}
\end{table*}

In Table~\ref{tab:devi}, we present the average of standard deviations of $L(\text{\pp})$, $E(\text{\p})$, and $\mathcal{C}(\text{\p})$ for reasoning models and non-reasoning models under different settings. The statistics show a high standard deviation because both $L(\text{\pp})$ and $E(\text{\p})$ have wide ranges, and it is not rare for them to reach extreme values: For $L(\text{\pp})$, considering components completely independently yields $L(\text{\pp})=U\cdot(N+M)=40$, while directly repeating the dataset yields $L(\text{\pp})=d\cdot(N+M)=320$; for $E(\text{\p})$, the value can be $0$ when completely correct and $d=16$ when completely incorrect.

In Table~\ref{tab:group}, we show the results of significance tests for cross-model performance comparisons. Overall, significant differences reveal a distinct gradient division.

\subsection{Qualitative Analysis}
\label{app:example}
Figure~\ref{fig:code} shows fragments of some of the programs generated by LLMs for qualitative analysis in Section~\ref{sec:qua}.

For Setting-Combination, LLM outputs can be categorized into two scenarios: (A) direct repetition of the dataset, a typical behavior under Random settings, and (B) being able to consider the Horizontal and Random components independently. The number of times scenario B occurred for reasoning models (out of 30 samples) is shown in Table~\ref{tab:times}. This ranking is identical to that of $L(\text{\pp})$, which serves as the basis for analysis (3) in Section~\ref{sec:qua}.

\begin{table}[h]
\centering
{
\begin{tabular}{l|c}
\hline
& \textbf{Times} \\
\hline
Claude-3.7	& 29 \\
o3-mini	& 27 \\
QwQ-Plus	& 20 \\
o1-mini	& 17 \\ 
DeepSeek-R1	& 15 \\
Gemini-2.5	& 13 \\
\hline
\end{tabular}
}
\caption{\label{tab:times} The number of times the reasoning model considers components of different setting independently under Setting-Combination.}
\end{table}

\subsection{Prompt Sensitivity Test}
\label{app:prompt}
We conduct experiments using three different styles of prompts under the Horizontal setting; the results are shown in Table~\ref{tab:sensi}. Despite fluctuations in the test results, these variations do not affect the final ranking of compositionality estimations.

\end{document}